%% file: main.tex
\documentclass[letterpaper,10pt,conference]{ieeeconf}
\IEEEoverridecommandlockouts
\input{preamble.tex}

\input{metadata.tex}

\DefineBibliographyStrings{english}{
    andothers = {\textit{et\addabbrvspace al\adddot}},
}

\AtBeginBibliography{\footnotesize\setlength{\bibitemsep}{0pt}}
\csdef{shortvenue:hu_motion_2024}{IROS}
\csdef{shortvenue:baril_subarctic_2020}{CRV}
\csdef{shortvenue:cariou_path_2010}{IROS}
\csdef{shortvenue:nayl_switching_2012}{MED}
\csdef{shortvenue:baril_drive_2024}{ICRA}
\csdef{shortvenue:sturm_benchmark_2012}{IROS}
\DeclareFieldFormat[inproceedings]{booktitle}{%
  \ifcsdef{shortvenue:\thefield{entrykey}}
    {\mkbibemph{\csuse{shortvenue:\thefield{entrykey}}}}
    {\mkbibemph{#1}}}

\graphicspath{{figures/}}

\begin{document}

\title{\LARGE\bfseries\PaperTitle\vspace{-3pt}}
\input{authors.tex}\IEEEaftertitletext{\vspace{-6pt}}

\maketitle

\begin{abstract}
\sisetup{mode=text,reset-text-series=false}
Trajectory planning and control in field robotics rely on predicting how propulsion and steering affect vehicle motion when contact points undergo slip.
For articulated vehicles, the \acf{PCK} accounts for the linkage geometry but neglects the rotational resistance distributed along the contacts.
We propose a \acf{FSQ} for single-track, center-articulated vehicles that incorporates this resistance through a quasi-static balance of lateral slip.
Our approach generalizes the standard \ac{PCK} formulation by relaxing the contact-point assumption.
An exact reduction of the quadratic slip cost to contact moments gives a compact closed-form solution for real-time prediction of lateral velocity and yaw rate.
We evaluate the proposed method in real-world experiments across asphalt, grass, ice, and mixed routes, using more than \SI{7}{\kilo\meter} of data.
For five-second predictions, \acs{FSQ} reduces the weighted median translation and yaw errors by \SI{53.5}{\percent} and \SI{68.9}{\percent}, respectively, relative to \acs{PCK}.
\end{abstract}

\section{Introduction}\label{sec:intro}

\begin{figure}[htbp]
  \centering
  \includegraphics[width=\columnwidth]{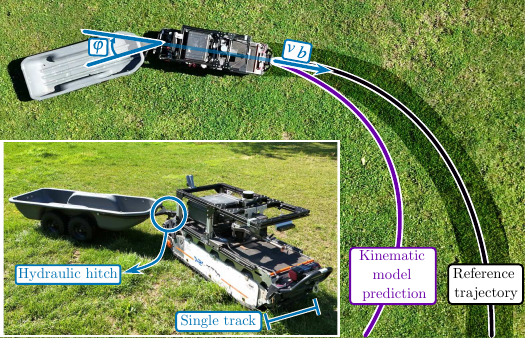}
  \caption{Articulated motion prediction during a field turn.
  The purple point-contact prediction, based on pointwise no-slip constraints, deviates markedly from the black traveled path.
  Blue annotations show longitudinal body velocity $v_b$ and articulation $\varphi$ in the overhead view, while the lower-left inset identifies the two actuation mechanisms: the driven track and hydraulic hitch.}
  \label{fig:hero}
  \vspace{-2pt}
\end{figure}

Planning, control, and state estimation in field robotics rely on motion models that relate propulsion and steering to changes in position and orientation~\citep{baril_subarctic_2020}.
Such models are needed when robots haul loads over snow or through narrow passages between trees.
These terrain and space constraints motivate a design that combines extended ground contact with a narrow powered unit.
Single-track articulated robots offer this combination and can tow interchangeable trailers or sleds suited to the load and terrain.
Unlike conventional skid-steered tracked vehicles, they steer through hitch articulation rather than the differential motion of two independently driven tracks.
In the configuration shown in \cref{fig:hero}, one driven front track pulls a passive four-wheel trailer through an actuated hydraulic hitch.
The tractor's position and heading, together with the articulation angle, define the vehicle's four planar degrees of freedom.
However, the robot has only two actuation channels: one for track propulsion and one for hitch articulation.
The vehicle is therefore underactuated, and the sideways motion and turning produced by these inputs depend on the ground reactions acting on both bodies.

A \acf{PCK} predicts this motion by imposing zero lateral velocity at one reference point on each body.
With vehicle dimensions as its only parameters, this model supports motion-primitive planning demonstrated on a full-scale articulated harvester~\citep{hu_motion_2024}.
Its simplicity comes from the no-slip assumption at those two points, whereas field experiments show how sliding on deformable or slippery terrain limits geometric motion prediction~\citep{baril_subarctic_2020}.
During a turn, a contact patch's front and rear move sideways even if its center has no lateral motion.
The resulting reactions on the tractor and trailer resist rotation and interact through the hitch.
Replacing each contact patch by one point retains the body-to-body lever arms, but omits the resistance to rotation produced across it.
The prediction task is therefore to recover both lateral velocity and yaw rate from these coupled contacts.

To account for this distributed resistance, our \acf{FSQ} allows both bodies to move sideways as they turn.
It predicts their motion by balancing lateral sliding over both ground-contact regions while neglecting inertia.
We show that the effect of these extended contacts can be retained in a compact motion predictor.
The model combines the vehicle and contact geometry with one parameter identified offline, which sets the relative influence of the front and rear contacts.
In the limit of vanishing contact extent around the reference points, the model recovers the complete point-contact baseline.
This connection also defines the comparison: both models receive the same realized longitudinal velocity, articulation angle, and articulation rate.
Their predicted trajectories are evaluated against recorded field motion.
Our contributions are
\begin{itemize}
  \setlength{\itemsep}{0pt}
  \setlength{\parskip}{0pt}
  \item a finite-support articulated motion model that couples front and rear slip through the hitch and reduces the distributed quadratic cost exactly to a $2\times2$ linear system,
  \item its analytical connection to complete point-contact kinematics, including the point-support limit and distinct steady and transient responses, and
  \item a field evaluation spanning over \SI{7}{\kilo\meter} on asphalt, grass, and ice, demonstrating the benefits of accounting for distributed contact resistance.
\end{itemize}

\section{Related Work}\label{sec:related}
Motion models represent ground contact at different levels, from rolling constraints at individual points to contact laws distributed over a wheel or track.
Under ideal rolling constraints, point-contact kinematics (\acs{PCK}) relates the motion of two articulated bodies by imposing zero lateral velocity at one reference point on each body~\citep{corke_steering_2001}.
The resulting steering law depends on vehicle dimensions, articulation angle, and articulation rate.
These same geometric relations provide a basis for path tracking, where a geometric criterion can regulate the lateral distances of both axle midpoints from a reference path~\citep{altafini_path-tracking_1999}.
A geometric state-space formulation similarly permits feedback on displacement, heading, and curvature errors for load-haul-dump vehicles~\citep{ridley_load_2003}.
Because the rolling constraints are imposed at points, these models retain the distances between the hitch and the reference points while omitting the extent of each contact patch.

Once sliding permits lateral motion at those points, the geometric prediction must be corrected to follow the observed path.
Off-road guidance can compensate for this deviation while anticipating curvature transitions, as demonstrated in full-scale agricultural path-tracking experiments~\citep{lenain_high_2006}.
For a vehicle--trailer system, estimated sideslip angles describe the departure from the rolling direction and allow the towing vehicle to correct the trailer's path~\citep{cariou_path_2010}.
To accommodate variations in slip on articulated vehicles, switching model predictive control uses models around different nominal slip angles, with performance evaluated in simulation~\citep{nayl_switching_2012}.
The slip description therefore supplies the correction needed for control, leaving the prediction of slip from contact properties as a separate modeling task.

To provide that physical explanation, terramechanics starts with the stresses developed at the ground.
For a towed wheel, it relates the force needed to pull it and its resulting skid to the normal and tangential stresses distributed over the wheel--soil contact~\citep{wong_towed_1967}.
A steady skid-steering theory similarly relates tracked-vehicle turning to the shear stress generated as the tracks slide over firm ground~\citep{wong_skid_2001}.
Experiments on natural snow further show that longitudinal tire response changes with slip, load, and the terrain disturbed by a preceding wheel~\citep{lee_slip-based_2012}.
Consequently, using these physical descriptions in a predictor requires soil and contact parameters as well as vehicle geometry.
Online estimation addresses this requirement for wheeled rovers by identifying soil cohesion and internal friction angle from onboard measurements~\citep{iagnemma_online_2004}.

When the objective is a compact motion predictor, an alternative is to identify the effect of slip directly from observed motion.
For skid-steering vehicles, identified equivalent track \acp{ICR} incorporate terrain-dependent slip into geometric prediction~\citep{martinez_approximating_2005}.
To learn this input--motion relation from data, an exploration protocol characterizes vehicle input limits and gathers training data for a learned slip model~\citep{baril_drive_2024}.
Building on this characterization, the extended protocol maps velocity commands to steady slip and compares commanded and realized motion to quantify command unpredictability~\citep{samson_drive_2025}.
For an articulated vehicle, the front and rear contact regions change orientation relative to each other as the joint moves.
The two bodies must assign the same velocity to their common hitch, but this condition alone does not determine their lateral velocities and yaw rates.
A mechanical model must therefore explain how the two contact responses combine to produce motion consistent with this geometric constraint.

To account for this inter-body coupling, mechanical steering models describe the ground reactions on each body and the forces transmitted at their connection.
An early steering analysis considers a tracked tractor and a tracked towed unit joined by a drawbar fixed to the towed unit~\citep{alhimdani_steering_1982}.
For articulated tracked vehicles, the influence of contact forces and load distribution has also been examined with a model compared against scale-model tests~\citep{watanabe_steerability_1986}.
A more recent steady steering model combines inter-unit coupling with load transfer and a shear-based contact description on firm ground~\citep{li_modeling_2025}.
Beyond steady steering, a nonlinear model with eight degrees of freedom includes planar motion of both units, track rotation, and hydraulic steering~\citep{tota_analytical_2021}.
Its linearization describes steady and transient cornering only at small lateral accelerations, and the analysis and hitch-angle controllers are tested in simulation.
However, using this model requires mass distribution, yaw inertias, cornering stiffnesses, and hydraulic parameters.
Here, we use realized longitudinal velocity and articulation history as inputs, so actuator dynamics need not be identified.
The remaining lateral velocity and yaw rate follow from a quasi-static contact balance, with one identified parameter setting the rear contact's effective resistance to lateral slip relative to the front. %

In this quasi-static setting, determining the remaining velocities requires a contact law that accounts for resistance to both translation and rotation.
For a finite contact, a limit surface describes the combinations of friction force and torque that it can sustain~\citep{goyal_planar_1991}.
For motion prediction, the power-dissipation method selects velocities that minimize frictional dissipation for prescribed actuation~\citep{murphey_power_2006}.
Applied to a rigid two-track vehicle, this method predicts motion over continuous or discrete contacts and stairs, improving on kinematic baselines in experiments~\citep{dixit_kinematics_2020}.
Its flat-ground closed form uses a fitted quadratic approximation of distributed Coulomb dissipation, but does not account for moving articulation.
Our \acs{FSQ} instead combines hitch compatibility with a quadratic lateral slip law for the front and rear contacts.
The quadratic law is a modeling assumption, while its reduction to contact moments is exact.
This reduction retains distributed rotational resistance without integrating over the contacts during prediction.
We use complete \acs{PCK}~\citep{corke_steering_2001} as the baseline because both models describe the same articulated vehicle using the same realized inputs, including articulation rate.
\acs{PCK} requires only the vehicle dimensions, while \acs{FSQ} adds the contact geometry and one identified weight ratio.
For centered contact weightings, vanishing contact extents recover \acs{PCK}, establishing a direct analytical connection between the two predictors.
This comparison therefore tests whether retaining finite contact extent improves the prediction of both position and heading over time.

\section{Theory}\label{sec:theory}

\begingroup
\setlength{\abovedisplayskip}{5pt plus 2pt minus 2pt}
\setlength{\belowdisplayskip}{5pt plus 2pt minus 2pt}
\setlength{\abovedisplayshortskip}{0pt plus 2pt}
\setlength{\belowdisplayshortskip}{5pt plus 2pt minus 2pt}

The proposed \acf{FSQ} predicts tractor lateral velocity $v_{y,1}$ and yaw rate $\omega_1$ from the realized longitudinal velocity $v_b$, articulation angle $\varphi$, and articulation rate $\dot\varphi$.
To relate the motion of the two bodies, we require both to assign the same velocity to their common hitch.
We first combine this compatibility with point-contact constraints to recover the \acf{PCK}, then replace these constraints with a distributed slip cost that reduces to a two-dimensional linear system.

\subsection{Geometry and hitch compatibility}\label{sec:platform}
To express how the hitch couples the two bodies, we use the geometry of the driven tractor ($i=1$, in teal) and passive-wheel trailer ($i=2$, in plum) shown in \cref{fig:frames}.
The body frames $\mathcal F_i$ originate at the nominal geometric centers $P_i$ of the front contact and rear axle group (right and left insets, respectively), with $x_i$ forward, $y_i$ lateral to the left, and $z_i$ normal to the plane.
The hitch $H$ lies a distance $\ell_1$ behind $P_1$ and $\ell_2$ ahead of $P_2$.
Headings $\theta_i$ are measured counterclockwise from the fixed world axis $x_W$, giving articulation $\varphi=\theta_2-\theta_1$.

\begin{figure}[htbp]
  \centering
  \includegraphics[width=\columnwidth]{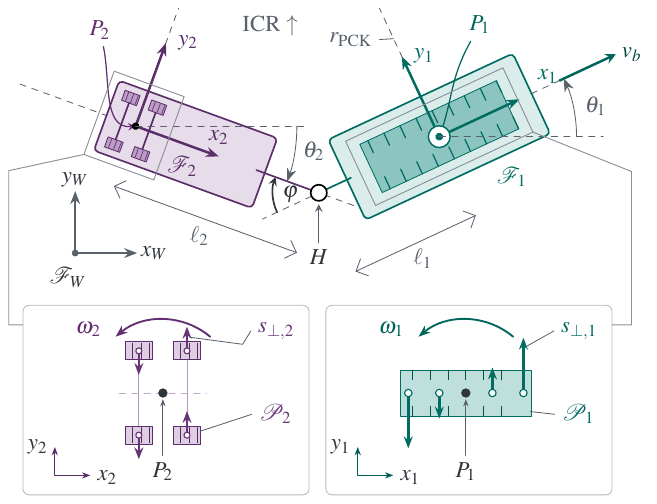}
  \caption{Articulated geometry and lateral contact slip.
The tractor is depicted in teal and the trailer in plum, with body frames $\mathcal F_i$ centered at $P_i$.
  Distances $\ell_i$ connect these origins to hitch $H$, $\varphi$ is articulation, and $v_b$ is tractor longitudinal velocity.
  Headings $\theta_i$ are measured from the world $x_W$ direction.
  Dashed lines perpendicular to $x_i$ lead to the off-frame \acf{ICR}, with turning radius $r_{\mathrm{PCK}}$ for steady point-contact motion.
  Gray guides locate the contact domains $\mathcal P_i$ enlarged below, with both inset frames oriented alike.
  In the insets, curved arrows indicate positive yaw rates $\omega_i$, and straight arrows show the contribution of rotation about $P_i$ to lateral slip $s_{\perp,i}$.}
  \label{fig:frames}
\end{figure}

Let $\mathbf T_{Wi}\in\SE(2)$ map body coordinates into the world frame.
The kinematic state $\mathbf X=(\mathbf T_{W1},\varphi)\in\SE(2)\times\Sph^1$ contains the tractor pose and the independent articulation coordinate.
The body twist collects the linear velocity of $P_i$ and the yaw rate of body $i$, relative to the world and expressed in $\mathcal F_i$:
\begin{equation}\label{eq:twist}
 \boldsymbol\xi_{Wi}
 =\begin{bmatrix}v_{x,i}&v_{y,i}&\omega_i\end{bmatrix}^{\!\top},
 \qquad
 \mathbf T_{Wi}^{-1}\dot{\mathbf T}_{Wi}
 =\boldsymbol\xi_{Wi}^{\wedge}.
\end{equation}
Here $\omega_i=\dot\theta_i$, and $(\cdot)^\wedge$ maps twist coordinates into the Lie algebra $\se(2)$~\citep{barfoot_state_2017}.
An overdot denotes a time derivative.
With the realized signed body velocity $v_{x,1}=v_b$ and articulation history $(\varphi,\dot\varphi)$ given, the model determines $v_{y,1}$ and $\omega_1$.

To relate the body twists through the hitch, we first express the relative pose using the lever arms in \cref{fig:frames}.
With $\ell_\varphi=\ell_1\cos\varphi+\ell_2$, the transform from tractor to trailer coordinates is
\begingroup
\setlength{\arraycolsep}{4pt}
\begin{equation}\label{eq:relative-pose}
 \mathbf T_{21}(\varphi)=\mathbf T_{W2}^{-1}\mathbf T_{W1}
 =\begin{bmatrix}
 \cos\varphi&\sin\varphi&\ell_\varphi\\
 -\sin\varphi&\cos\varphi&-\ell_1\sin\varphi\\
 0&0&1
 \end{bmatrix}.
\end{equation}
\endgroup
Differentiating $\mathbf T_{W2}=\mathbf T_{W1}\mathbf T_{21}^{-1}$ then expresses the trailer twist in terms of the tractor twist and the changing articulation:
\begin{equation}\label{eq:adjoint}
\begin{aligned}
 \boldsymbol\xi_{W2}&=\Ad(\mathbf T_{21})\boldsymbol\xi_{W1}
                         +\boldsymbol\xi_{12},\\
 \boldsymbol\xi_{12}^{\wedge}&=-\dot{\mathbf T}_{21}\mathbf T_{21}^{-1}.
\end{aligned}
\end{equation}
Here $\Ad(\mathbf T_{21})$ transports the tractor twist to the origin and axes of $\mathcal F_2$~\citep{murray_mathematical_1994}.
For the hitch geometry in \cref{fig:frames}, the relative twist expressed in $\mathcal F_2$ is
\begin{equation}\label{eq:relative-twist}
 \boldsymbol\xi_{12}=\dot\varphi\begin{bmatrix}0&-\ell_2&1\end{bmatrix}^{\!\top}.
\end{equation}
Its lateral term $-\ell_2\dot\varphi$ is induced at $P_2$ by rotation about the hitch.
Hitch compatibility therefore determines the trailer twist once $v_{y,1}$ and $\omega_1$ are known.
A contact law is still needed to determine these two tractor velocities.

\subsection{Point-contact kinematics}\label{sec:hierarchy}
Using this hitch relation, the \acf{PCK} determines motion by imposing zero lateral velocity at both reference points~\citep{corke_steering_2001}.
To apply these constraints, we use the lateral and yaw components of \cref{eq:adjoint}:
\begin{equation}\label{eq:hitch-components}
\begin{aligned}
 v_{y,2}&=\cos\varphi\,v_{y,1}-\ell_\varphi\omega_1
              -v_b\sin\varphi-\ell_2\dot\varphi,\\
 \omega_2&=\omega_1+\dot\varphi.
\end{aligned}
\end{equation}
For the \acs{PCK} baseline, setting $v_{y,1}=v_{y,2}=0$ in \cref{eq:hitch-components} gives
\begin{equation}\label{eq:m0}
 v_{y,1}=0,\qquad
 \omega_1=-\frac{v_b\sin\varphi+\ell_2\dot\varphi}{\ell_\varphi},
 \quad \ell_\varphi\ne0.
\end{equation}
Kinematically, the numerator separates advance at fixed articulation from joint rotation, which can produce yaw even at $v_b=0$.
For forward steady turns with $|\varphi|<\pi/2$, positive articulation therefore produces negative tractor yaw.
At fixed articulation and nonzero $v_b\sin\varphi$, the turning radius in \cref{fig:frames} is $r_{\mathrm{PCK}}=|\ell_\varphi/\sin\varphi|$.
This geometric model retains the hitch lever arms but no contact extent.

\subsection{Finite-support quadratic closure}\label{sec:dissipative}\label{sec:reduction}
\acs{PCK}'s point constraints omit slip away from the contact centers.
During rotation, a contact patch's front and rear slip laterally in opposite directions even when its center has no lateral velocity, as illustrated in \cref{fig:frames}.
Our \acs{FSQ} model allows center motion to balance this distributed slip.
We assume rigid bodies and quasi-static contact with fixed weightings, neglecting inertia and slip memory.
Let $\mathcal P_i\subset\R^2$ denote body $i$'s contact region, with coordinates $(x,y)$ measured from $P_i$ in $\mathcal F_i$.
At longitudinal coordinate $x$, the lateral slip velocity $s_{\perp,i}(x)$ is the contact material's velocity relative to stationary ground along $y_i$:
\begin{equation}\label{eq:slip}
 s_{\perp,i}(x)=v_{y,i}+\omega_i x.
\end{equation}
The center velocity $v_{y,i}$ adds uniform lateral motion, while the rotational term $\omega_i x$ varies linearly and reverses sign across $P_i$.

To account for resistance over each contact, \acs{FSQ} penalizes the squared lateral slip in \cref{eq:slip}.
This treats both slip directions equally and permits evaluation from contact moments.
For the front track, we use a uniformly weighted rectangle $\mathcal P_1$ of length $l_t>0$, centered at $P_1$ and aligned with $\mathcal F_1$.
The rear contact comprises two equally weighted axles separated longitudinally by $d_r$, at $x_{2q}=\pm d_r/2$ relative to $P_2$, with $c_{2q}=c_2/2$.
The unknown coefficients $c_i>0$ set each body's total effective resistance weight.
Since $s_{\perp,1}(x)$ is independent of $y$, integration across the track width cancels that width from the area normalization.
The front and rear contact costs are
\begin{equation}\label{eq:dissipation}
\begin{aligned}
 d_1(\boldsymbol\xi_{W1})
 &=\frac{c_1}{2|\mathcal P_1|}
   \int_{\mathcal P_1}s_{\perp,1}(x)^2\,\dd x\,\dd y\\
 &=\frac{c_1}{2l_t}
   \int_{-l_t/2}^{l_t/2}(v_{y,1}+\omega_1x)^2\,\dd x,\\
d_2(\boldsymbol\xi_{W2})
 &=\frac12\sum_q c_{2q}s_{\perp,2}(x_{2q})^2\\
 &=\frac{c_2}{4}\left[
   \left(v_{y,2}-\frac{d_r}{2}\omega_2\right)^2+
   \left(v_{y,2}+\frac{d_r}{2}\omega_2\right)^2
   \right].
\end{aligned}
\end{equation}
Here $x_{2q}$ and $c_{2q}\ge0$ denote the longitudinal position and weight of rear axle $q$, with total weight $c_2=\sum_q c_{2q}>0$.
In $d_2$, we express $v_{y,2}$ and $\omega_2$ in terms of the tractor velocities and prescribed inputs using the hitch relations in \cref{eq:hitch-components}.

Each contact cost is independent of $v_{x,i}$ and therefore cannot recover the full body twist on its own.
With $v_{x,1}=v_b$ supplied, we minimize over $\mathbf y=(v_{y,1},\omega_1)^\top$, while \cref{eq:adjoint} determines the trailer twist.
The tractor twist is
\begin{equation}\label{eq:free-velocities}
\begin{aligned}
 \boldsymbol\xi_{W1}&=\boldsymbol\xi^\circ+\mathbf B\mathbf y,
 &\boldsymbol\xi^\circ&=\begin{bmatrix}v_b\\0\\0\end{bmatrix},
 &\mathbf B&=\begin{bmatrix}0&0\\1&0\\0&1\end{bmatrix}.
\end{aligned}
\end{equation}
Here $\boldsymbol\xi^\circ$ contains the prescribed advance, and $\mathbf B$ inserts the two unknown velocities.
The individual contact weights $c_i$ cannot be identified from motion alone, since a common positive scaling leaves the minimizing velocities unchanged.
We therefore divide the combined cost by $c_1$ and retain only the ratio $\alpha=c_2/c_1>0$.
We learn this ratio from experimental data rather than estimating the individual weights, as described in \cref{sec:identification}.
Using the twists from \cref{eq:free-velocities,eq:adjoint}, the normalized cost is
\begin{equation}\label{eq:combined-cost}
 j_\alpha(\mathbf y)=\frac{d_1(\boldsymbol\xi_{W1}(\mathbf y))}{c_1}
                   +\alpha\frac{d_2(\boldsymbol\xi_{W2}(\mathbf y))}{c_2}.
\end{equation}
Following quasi-static dissipation minimization~\citep{murphey_power_2006}, we use this quadratic cost to select the velocities compatible with the prescribed inputs:
\begin{equation}\label{eq:argmin}
 \mathbf y^\star=\argmin_{\mathbf y\in\R^2}j_\alpha(\mathbf y).
\end{equation}
Increasing $\alpha$ gives greater weight to rear slip.
At $\alpha=1$ the contacts have equal total weight, and as $\alpha$ tends to zero the rear contribution vanishes.

To avoid contact integration during prediction, we use the longitudinal centroid offset $a_i$ and squared spread $\rho_i^2$.
These are the weighted means of $x$ and $(x-a_i)^2$, respectively.
Expanding the squared slip in either the continuous or discrete cost gives
\begin{equation}\label{eq:moment-identity}
 d_i(\boldsymbol\xi_{Wi})
 =\frac{c_i}{2}\Bigl[(v_{y,i}+a_i\omega_i)^2+\rho_i^2\omega_i^2\Bigr].
\end{equation}
For the centered front track and equally weighted rear axle pair, the contact moments are
\begin{equation}\label{eq:moments}
 a_1=a_2=0,\qquad
 \rho_1^2=\frac{l_t^2}{12},\qquad
 \rho_2^2=\frac{d_r^2}{4}.
\end{equation}
We retain $a_i$ below so the derivation also covers noncentered contact weightings.
In \cref{eq:moment-identity}, the first term penalizes lateral slip at the weighted centroid, while the second retains the rotational contribution illustrated in \cref{fig:frames}.
A zero spread removes the second term, while a larger spread increases the rotational penalty at fixed total weight.
For \acs{FSQ}, we assume at least one positive contact spread, which ensures a unique solution to \cref{eq:argmin}.

To assemble both contacts in one linear system, we write \cref{eq:moment-identity} as a quadratic form in the body twist.
Normalizing by $c_1$ gives the contact matrix $\mathbf Q_i$:
\begin{equation}\label{eq:contact-quadratic-form}
\begin{aligned}
 d_i(\boldsymbol\xi)/c_1&=\tfrac12\boldsymbol\xi^\top\mathbf Q_i\boldsymbol\xi,\\
 \mathbf Q_i&=\frac{c_i}{c_1}\begin{bmatrix}
 0&0&0\\
 0&1&a_i\\
 0&a_i&a_i^2+\rho_i^2
 \end{bmatrix}.
\end{aligned}
\end{equation}
Here $\mathbf Q_2$ includes $\alpha$, while the zero first row and column reflect the restriction to lateral slip.
To combine the contact costs, we use \cref{eq:free-velocities,eq:adjoint} to express both twists in the same unknowns $\mathbf y$.
The Hessian matrix $\mathbf H$ describes the coupled penalty on lateral velocity and yaw.
With $\mathbf g$ denoting the gradient of $j_\alpha$ at $\mathbf y=\mathbf0$ and $\mathrm{const}$ collecting terms independent of $\mathbf y$, the cost is
\begin{equation}\label{eq:free-quadratic}
 j_\alpha(\mathbf y)=\tfrac12\mathbf y^\top\mathbf H\mathbf y
+\mathbf g^\top\mathbf y+\mathrm{const}.
\end{equation}
To obtain the two velocities, we differentiate \cref{eq:free-quadratic} with respect to $\mathbf y$ and set the gradient to zero:
\begin{equation}\label{eq:contact-balance}
 \nabla_{\mathbf y}j_\alpha(\mathbf y)=\mathbf H\mathbf y+\mathbf g=\mathbf0.
\end{equation}
Collecting the quadratic and linear terms in $\mathbf y$ gives the coefficients needed to solve \cref{eq:contact-balance}:
\begin{equation}\label{eq:lie-reduced-system}
\begin{aligned}
 \mathbf H&=\mathbf B^\top\!\left(\mathbf Q_1+
       \Ad(\mathbf T_{21})^\top\mathbf Q_2\Ad(\mathbf T_{21})\right)\mathbf B,\\
 \mathbf g&=\mathbf B^\top\Ad(\mathbf T_{21})^\top\mathbf Q_2
       (\Ad(\mathbf T_{21})\boldsymbol\xi^\circ+\boldsymbol\xi_{12}).
\end{aligned}
\end{equation}
Since $\mathbf Q_1\boldsymbol\xi^\circ=\mathbf0$, the prescribed motion contributes to $\mathbf g$ only through the rear contact.

To show that this balance determines a unique motion, we write the quadratic form of $\mathbf H$ as a sum of squares.
For a velocity variation $\delta\mathbf y=(\delta v_{y,1},\delta\omega_1)^\top$,
\begin{equation}\label{eq:gram-positivity}
\begin{split}
 \delta\mathbf y^\top\mathbf H\delta\mathbf y
 ={}&(\delta v_{y,1}+a_1\delta\omega_1)^2\\
 &+\alpha[\cos\varphi\,\delta v_{y,1}+(a_2-\ell_\varphi)\delta\omega_1]^2\\
 &+(\rho_1^2+\alpha\rho_2^2)\delta\omega_1^2.
\end{split}
\end{equation}
A positive spread forces $\delta\omega_1=0$ and then $\delta v_{y,1}=0$ whenever this sum vanishes.
Hence $\mathbf H$ is positive definite even for noncentered weightings, and the unique solution of \cref{eq:argmin} is
\begin{equation}\label{eq:velocity-solve}
 \mathbf y^\star=-\mathbf H^{-1}\mathbf g.
\end{equation}
With the contact moments fixed and $\alpha$ identified, \cref{eq:lie-reduced-system,eq:velocity-solve} give both tractor velocities at each input sample.
The trailer twist then follows from \cref{eq:adjoint}.

To predict the vehicle pose over time, we insert this velocity solution into \cref{eq:free-velocities,eq:twist}, giving the state dynamics:
\begin{equation}\label{eq:state-model}
\begin{aligned}
 \dot{\mathbf X}&=\left(\mathbf T_{W1}
       (\boldsymbol\xi^\circ+\mathbf B\mathbf y^\star)^\wedge,\,\dot\varphi\right),\\
 \mathbf T_{W2}&=\mathbf T_{W1}\mathbf T_{21}(\varphi)^{-1}.
\end{aligned}
\end{equation}
Starting from $\mathbf T_{W1,0}$, we integrate on $\SE(2)$ by composing increments from the mean twist between consecutive samples.
For intervals $\Delta t_k=t_{k+1}-t_k$, the pose after $n$ steps is
\begin{equation}\label{eq:pose-update}
\begin{aligned}
 \overline{\boldsymbol\xi}_{W1,k}
 &=\tfrac12\bigl(\boldsymbol\xi_{W1,k}+\boldsymbol\xi_{W1,k+1}\bigr),\\
 \mathbf T_{W1,n}
 &=\mathbf T_{W1,0}\prod_{k=0}^{n-1}
       \exp\!\left(\Delta t_k\,\overline{\boldsymbol\xi}_{W1,k}^{\wedge}\right).
\end{aligned}
\end{equation}
Because we use body twists, increments multiply on the right in chronological order~\citep{barfoot_state_2017}.
The updated tractor pose and supplied articulation then determine the trailer pose through \cref{eq:state-model}.
Using \cref{eq:velocity-solve} for \acs{FSQ} or \cref{eq:m0} for \acs{PCK} advances the predicted poses with the same update in either travel direction.

\subsection{Connection to point-contact prediction}\label{sec:steering}\label{sec:implementation}
We first establish how \acs{FSQ} recovers \acs{PCK} as the modeled contact extents tend to zero, then explain how finite contact changes the predicted yaw.
The limit removes the positive-spread assumption in \cref{eq:gram-positivity}, so we check that $\mathbf H$ remains invertible.
Setting $a_1=a_2=0$ in \cref{eq:lie-reduced-system} gives its determinant,
\begin{equation}\label{eq:determinant}
 \det\mathbf H=\alpha\ell_\varphi^2+(1+\alpha\cos^2\varphi)(\rho_1^2+\alpha\rho_2^2).
\end{equation}
As both spreads vanish, $\det\mathbf H\to\alpha\ell_\varphi^2>0$ for fixed $\alpha>0$ and $\ell_\varphi\ne0$.
The limiting cost in \cref{eq:moment-identity} is minimized by $v_{y,1}=v_{y,2}=0$, so the velocities converge to the complete \acs{PCK} law in \cref{eq:m0}, including its articulation-rate term.

To see how finite contact extent modifies the predicted yaw rate, we compare \acs{FSQ} and \acs{PCK} under the same measured inputs, first at fixed articulation and then during joint motion.
For centered contacts with at least one positive spread and $\ell_\varphi\ne0$, we eliminate $v_{y,1}$ from \cref{eq:contact-balance}.
At fixed articulation ($\dot\varphi=0$), the resulting yaw rate is the \acs{PCK} rate in \cref{eq:m0} multiplied by the dimensionless scalar gain
\begin{equation}\label{eq:steering-gain}
 \gamma(\varphi)=\frac{\alpha\ell_\varphi^2}{\det\mathbf H},
 \qquad 0<\gamma(\varphi)<1.
\end{equation}
In a steady turn, \acs{PCK} therefore predicts a larger yaw-rate magnitude than \acs{FSQ} in either travel direction, while $\gamma(\varphi)\to1$ as the modeled contact extents tend to zero.
When articulation changes, the rear rotational penalty in \cref{eq:moment-identity} involves $\omega_2=\omega_1+\dot\varphi$ from \cref{eq:hitch-components}.
Repeating the elimination while retaining $\dot\varphi$ gives the \acs{PCK} rate scaled by $\gamma(\varphi)$, plus a correction due to rear contact extent:
\begin{equation}\label{eq:transient-yaw}
\begin{split}
 \omega_1={}&-\gamma(\varphi)\frac{v_b\sin\varphi+\ell_2\dot\varphi}{\ell_\varphi}\\
 &-\frac{\alpha(1+\alpha\cos^2\varphi)\rho_2^2}{\det\mathbf H}\dot\varphi.
\end{split}
\end{equation}
\acs{FSQ} therefore changes both the yaw at fixed articulation and its response to joint motion, rather than applying a constant correction to \acs{PCK}.

\par
\endgroup

\section{Methodology}\label{sec:methodology}
To test whether finite contact extent improves trajectory prediction, we compare \acs{PCK} and \acs{FSQ} using the same realized inputs and initial poses.

\subsection{Platform, reference, and common inputs}
\begin{figure}[htbp]
 \centering
 \includegraphics[width=\columnwidth]{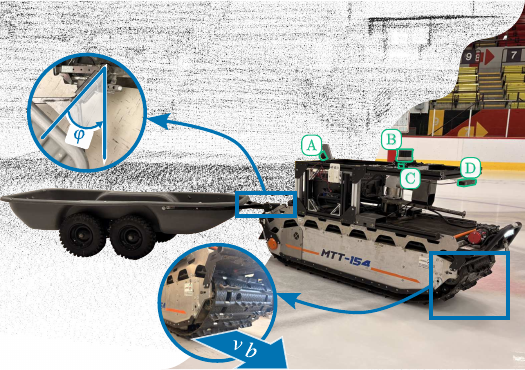}
 \caption{Field platform and instrumentation over a lidar map.
 Green labels identify \textbf{(A)} the RS Airy lidar, \textbf{(B)} the Hesai XT-32 lidar, \textbf{(C)} the Xsens MTi-100 \acf{IMU}, and \textbf{(D)} the ZED 2i camera.
 \textbf{A} provides articulation and trailer pose estimates in $\SE(3)$, while measurements from \textbf{B}, \textbf{C}, and \textbf{D} are combined to estimate tractor odometry.
 Blue insets locate the driven track and hydraulic hitch, with body velocity $v_b$ and articulation angle~$\varphi$.}
 \label{fig:setup}
\end{figure}

We use the platform in \cref{fig:setup} with the measured geometry in \cref{tab:experimental}.
For its uniformly weighted front track and equally weighted rear axles, we use the centered contact moments derived in \cref{sec:dissipative}.
Only the relative contact weight $\alpha$ is fitted from motion.

To construct the motion reference, we first estimate lidar poses offline by \ac{ICP} registration using libpointmatcher~\citep{pomerleau_comparing_2013}.
A factor-graph smoother combines these poses, inertial measurements, and visual odometry to recover tractor motion at \SI{100}{\hertz}.
Rigid sensor transforms express reference poses and velocities at $P_1$, including the rotational lever-arm contribution.
In the main comparison, both predictors receive the realized longitudinal body velocity $v_b$ from this reference and the encoder angle $\varphi$.
To obtain $\dot\varphi$, a centered third-order Savitzky--Golay filter differentiates $\varphi$ offline over a \SI{0.21}{\second} window without crossing timestamp gaps.

\begin{table}[htbp]
\vspace*{5pt}
\centering%
\caption{Platform geometry and contact parameters.
Only the rear/front contact-weight ratio is fitted from motion data.}
\sisetup{round-mode=places,round-precision=2}
\setlength{\tabcolsep}{3pt}
\renewcommand{\arraystretch}{1.08}
\begin{tabularx}{\columnwidth}{@{}c >{\raggedright\arraybackslash}X c r c@{}}
\toprule
 & \textbf{Characteristic} & \textbf{Symbol} & \textbf{Value} & \textbf{Unit}\\
\midrule
\multirow{6}{*}{\rotatebox[origin=c]{90}{\textbf{Robot}}}
 & Tractor mass & $m_1$ & \num[round-mode=none]{365} & \si{\kilo\gram}\\
 & Trailer mass & $m_2$ & \num[round-mode=none]{75} & \si{\kilo\gram}\\
 & Front center--hitch distance & $\ell_1$ & \num{0.8342} & \si{\meter}\\
 & Hitch--rear center distance & $\ell_2$ & \num{1.5145} & \si{\meter}\\
 & Front contact length & $l_t$ & \num{1.225} & \si{\meter}\\
 & Rear axle separation & $d_r$ & \num{0.455} & \si{\meter}\\
\midrule
\multirow{3}{*}{\rotatebox[origin=c]{90}{\textbf{Contact}}}
 & Front spread & $\rho_1$ & \num{0.3536} & \si{\meter}\\
 & Rear spread & $\rho_2$ & \num{0.2275} & \si{\meter}\\
 & Rear/front contact weight & $\alpha$ & $\simeq$ \num{0.20} & ---\\
\bottomrule
\end{tabularx}
\label{tab:experimental}
\end{table}

\subsection{Prediction segments}
The main comparison uses trajectory segments from nine field runs: three on asphalt, two on grass, two on ice, and two on mixed routes alternating between asphalt and grass around campus.
We select usable data independently of prediction scores, checking reference validity, initialization confidence, and articulation feedback.
Linear input interpolation bridges excluded stretches up to \SI{0.5}{\second}, affecting \num{288} five-second segments.
Longer exclusions and timestamp gaps interrupt prediction.
We extract disjoint segments separately at \num{100} horizons from \num{0.1} to \SI{10}{\second}, requiring at least \SI{0.2}{\meter} of travel per segment.
The segment population therefore varies with horizon and includes forward, reverse, and changing-direction motion.
The selection yields \num{1690} five-second segments and \num{992} ten-second segments, the latter covering \SI{7.46}{\kilo\meter} when each segment's reference travel is counted once.
Within each segment, both models start at the same reference pose and use the $\SE(2)$ update in \cref{eq:pose-update}, which averages the twists of consecutive samples (nominally \SI{0.01}{\second} apart), without subsequent pose correction.

\subsection{Contact identification}\label{sec:identification}
Before prediction, we fit $\alpha$ to reference lateral velocity and yaw rate from three of the nine runs, one each on grass, ice, and asphalt.
Multiplying yaw-rate residuals by $\ell_1+\ell_2$ puts both residuals on a common velocity scale.
To prevent densely sampled maneuvers from dominating the fit, we group straight motion, steady turns, and articulation changes by travel direction.
Weights then balance runs, groups within each run, their segments, and observations.
We apply a soft-$L_1$ loss with scale \SI{0.10}{\meter\per\second} to the square-root-weighted residuals and optimize over $\ln\alpha\in[-4,1]$.
The resulting $\alpha\simeq0.20$ remains fixed in the main comparison and campus examples.

To assess whether terrain-specific weights improve prediction, we also fit a separate ratio on each terrain's evaluated runs.
To test both calibrations on other runs, we fit one global and four terrain-specific ratios to \num{19780} observations from the three calibration runs plus one mixed-route run, with the same objective and bounds.
The five remaining runs provide \num{567} five-second segments, evaluated with the global and matching terrain ratios.
All ratios remain fixed during prediction.

\section{Results and Discussion}\label{sec:protocol}
We compare predicted and reference poses at the tractor's front contact center $P_1$ using translation and yaw \acp{RPE}~\citep{sturm_benchmark_2012}.
With the common initial pose, translation \acs{RPE} is the endpoint distance, and yaw \acs{RPE} is the absolute wrapped heading difference.

\subsection{Prediction error over the horizon}\label{sec:endpoint-rollout}
\begin{figure}[htbp]
 \centering
 \includegraphics[width=\columnwidth]{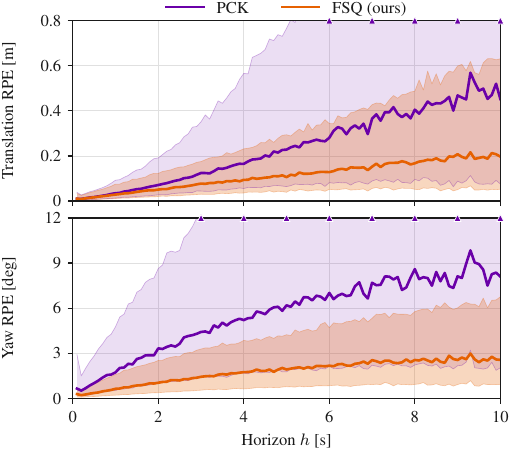}
 \caption{Prediction error versus horizon.
 Lines and bands show weighted medians and the \acf{IQR} of translation \acf{RPE} at $P_1$ (top) and tractor yaw \acs{RPE} (bottom), with equal weight per run.
 Triangles mark bands extending beyond the view.}
 \label{fig:horizon_rpe_evaluation}
 \vspace{-4pt}
\end{figure}

We first compare prediction errors over horizons $h$ from \SI{0.1}{\second} to \SI{10}{\second} in \cref{fig:horizon_rpe_evaluation}.
To give runs equal influence within each horizon or terrain group, we weight segments inversely to their number per run.

\acs{FSQ} yields lower weighted medians over most horizons.
At five seconds, the translation median falls from \SI{0.23}{\meter} for \acs{PCK} to \SI{0.11}{\meter} for \acs{FSQ}, while yaw falls from \SI{6.2}{\degree} to \SI{1.9}{\degree}.
These correspond to reductions of \SI{53.5}{\percent} and \SI{68.9}{\percent} in the weighted median levels.
Both five-second median errors decrease in all nine runs, including the six not used to fit the global ratio.
At ten seconds, the translation upper quartile decreases from \SI{2.1}{\meter} to \SI{0.63}{\meter}.
These results support retaining contact extent for multi-second prediction without pose correction.
The benefit is not confined to the calibration runs, and the lower translation upper quartile shows that it is not limited to median errors.

\subsection{Variation across recorded environments}
\begin{figure}[htbp]
 \centering
 \includegraphics[width=\columnwidth]{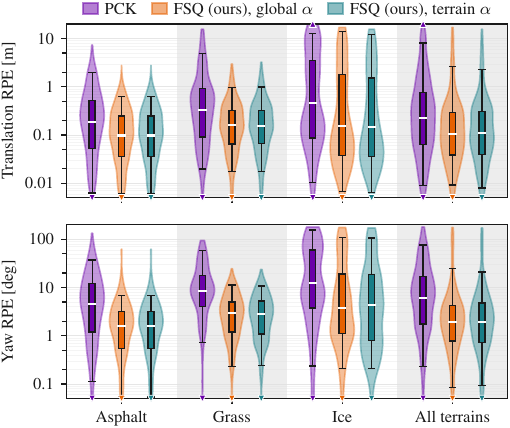}
 \caption{Five-second prediction errors and contact calibration.
 The violins show translation \acs{RPE} (top) and yaw \acs{RPE} (bottom) on logarithmic axes, with width representing error density.
 White marks, boxes, and whiskers indicate weighted medians, \acsp{IQR}, and fifth to 95\textsuperscript{th} percentiles, with equal total weight per run within each group.
 Triangles mark individual segment errors outside the view.
 \acs{FSQ} uses a global rear/front contact-weight ratio $\alpha$ or a separate ratio fitted on each terrain's evaluated runs.}
 \label{fig:terrain_boxplot}
 \vspace{-4pt}
\end{figure}

To check whether the overall improvement holds across ground conditions, \cref{fig:terrain_boxplot} separates the five-second prediction errors by environment.
With a global ratio, \acs{FSQ} lowers both weighted median errors relative to \acs{PCK} in every displayed group.
Large errors nevertheless remain on ice, with a five-second translation upper quartile of \SI{1.8}{\meter} for \acs{FSQ} versus \SI{3.5}{\meter} for \acs{PCK}.
In one ice-rink run, ten-second translation medians reach \SI{7.5}{\meter} for \acs{PCK} and \SI{3.0}{\meter} for \acs{FSQ}.

To assess whether these residuals call for terrain-specific weights, \cref{fig:terrain_boxplot} also compares global and per-terrain calibration.
Separate ratios have little effect on the asphalt and grass medians, while on ice the translation median decreases slightly and the yaw median increases.
To check whether these findings extend beyond the fitting runs, we evaluate both calibrations on five runs excluded from fitting, using ratios identified on the other four.
The global fit again gives $\alpha\simeq0.20$.
Global and per-terrain calibration yield similar five-second weighted medians: \SI{0.13}{\meter} and \SI{0.14}{\meter}, respectively, in translation, and approximately \SI{2.1}{\degree} in yaw for both.
The similarity of the errors on runs excluded from fitting supports using one global contact-weight ratio across the evaluated environments.
Terrain-specific calibration provides no consistent additional benefit and does not remove the substantial errors on ice.

\subsection{Prediction across longitudinal velocities}
\begin{figure}[htbp]
 \vspace*{5pt}
 \centering
 \includegraphics[width=\columnwidth]{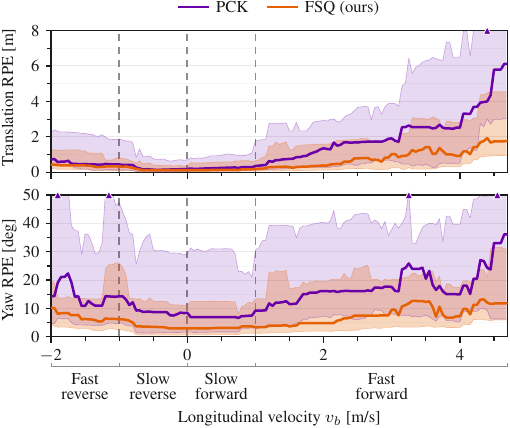}
 \caption{Five-second prediction errors grouped by longitudinal velocity.
 At each plotted velocity, lines and bands summarize the weighted medians and \acsp{IQR} of segments whose starting velocity lies within \SI{\pm0.75}{\meter\per\second} of that value, with equal total weight per run in that neighborhood.
 Direction and speed may change during prediction.
 Dashed lines mark zero and $|v_b|=\SI{1}{\meter\per\second}$, while triangles indicate \acsp{IQR} extending beyond the vertical limits.}
 \label{fig:velocity-rpe}
 \vspace{-10pt}
\end{figure}

To assess how prediction accuracy varies with speed and direction, we add \num{433} five-second segments from \num{12} runs with lidar-derived articulation.
\cref{fig:velocity-rpe} groups all \num{2123} segments from \num{21} runs by their starting longitudinal velocity.
Both models receive the same realized inputs, and \acs{FSQ} uses the fixed global contact ratio.

Within the slow region marked by the dashed lines ($|v_b|<\SI{1}{\meter\per\second}$), the translation medians are close, whereas the yaw medians remain lower for \acs{FSQ}.
In the neighborhood centered at \SI{-2}{\meter\per\second}, translation decreases from \SI{0.73}{\meter} for \acs{PCK} to \SI{0.43}{\meter} for \acs{FSQ}, while yaw decreases from \SI{14}{\degree} to \SI{10}{\degree}.
At \SI{2}{\meter\per\second}, the corresponding medians fall from \SI{1.4}{\meter} to \SI{0.39}{\meter} and from \SI{16}{\degree} to \SI{4.9}{\degree}.
Some jumps in the curves also reflect changes in which maneuvers and runs contribute to each velocity neighborhood.
These gains are obtained with the same contact calibration in both travel directions.
The translation benefit is clearest for segments starting at higher speeds, whereas the lower-speed groups mainly show an improvement in yaw.

\subsection{Interpretation and limits}\label{sec:trajectory-interpretation}
\begin{figure}[!htbp]
 \vspace*{5pt}
 \centering
 \includegraphics[width=\columnwidth]{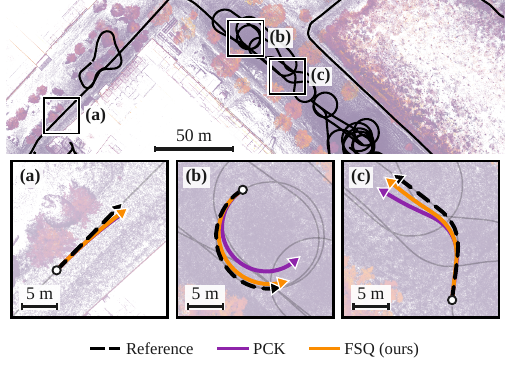}
 \caption{Campus trajectory with three five-second prediction examples.
 Boxes on the mapped reference route locate (a) a near-straight segment, (b) a sharp turn, (c) a milder turn. Background point colors indicate height.
 Insets use a common metric scale, with dashed black for the reference, purple for the point-contact model (\acs{PCK}), and orange for the \acf{FSQ}.
 Hollow circles mark the common starts and arrowheads the endpoints.
 Scale bars show \SI{50}{\meter} above and \SI{5}{\meter} below.}
 \label{fig:campus-multiview}
 \vspace{-8pt}
\end{figure}

Three five-second predictions from a separate campus run in \cref{fig:campus-multiview} illustrate these differences along individual paths.

The near-straight segment in panel~(a) gives similar endpoint translation errors, \SI{1.0}{\meter} for \acs{PCK} and \SI{0.89}{\meter} for \acs{FSQ}.
In panel~(b), \acs{PCK} cuts inside the \ang{157} turn, giving \SI{4.8}{\meter} of error versus \SI{1.3}{\meter} for \acs{FSQ}.
The milder \ang{57} turn in panel~(c) gives \SI{2.9}{\meter} for \acs{PCK} and \SI{1.3}{\meter} for \acs{FSQ}.

In these examples, the improvement is more pronounced in turns than in near-straight travel.
The wider turn predicted by \acs{FSQ} in \cref{fig:campus-multiview}(b) is consistent with the rotational resistance retained in \cref{eq:moment-identity}.
Relative to \acs{PCK}, finite contact extent reduces the yaw-rate magnitude at fixed articulation and changes the response to joint motion, as described by \cref{eq:steering-gain,eq:transient-yaw}.
These changes affect the path together with lateral slip at the contact centers, since yaw redirects subsequent travel while slip displaces the vehicle sideways.
Through \cref{eq:state-model}, these differences also carry over to the predicted trailer pose, which matters in narrow passages but is not evaluated here.

Despite these gains, we hypothesize that neglected inertial effects contribute to the large residual errors on ice.
Changing loads and reference or input errors may also contribute, but their effects are not separated here.
Extending the model to include these effects is left for future work.

\section{Conclusion}\label{sec:conclusion}

In this paper, we present a \acf{FSQ} that extends the \acf{PCK} for single-track center-articulated vehicles.
An exact reduction to contact moments yields lateral velocity and yaw rate from a compact system, retaining distributed rotational resistance with a single identified rear/front weight ratio.
We show that the model generalizes \acs{PCK} by recovering its complete equations as centered contact extents vanish.
Retaining their extent instead reduces the steady-turn yaw-rate magnitude relative to \acs{PCK} and changes the response to joint motion.

We evaluate our method on asphalt, grass, ice, and mixed routes using measured inputs in forward, reverse, and changing-direction motion.
The results show that \acs{FSQ} predicts the vehicle motion more accurately than \acs{PCK} across the evaluated conditions.

Future work will develop control laws based on this model and extend it to account for vehicle dynamics.
We will also investigate whether adapting the contact-weight ratio improves prediction when the payload changes.

\input{acknowledgments.tex}

\printbibliography

\end{document}

%% file: preamble.tex
\usepackage{amsmath,amssymb,amsfonts}
\usepackage{mathtools}
\usepackage{bm}
\usepackage{graphicx}
\usepackage{flafter}
\usepackage{placeins}
\usepackage{textcomp}
\usepackage{microtype}
\usepackage{needspace}
\usepackage{xcolor}
\usepackage{booktabs}
\usepackage{multirow}
\usepackage{silence}
\usepackage{xspace}

\usepackage[
  unicode=true,
  pdfencoding=auto,
  colorlinks=true,
  hypertexnames=false,
  linkcolor=blue!55!black,
  citecolor=blue!55!black,
  urlcolor=blue!55!black
]{hyperref}
\usepackage{hyperxmp}
\usepackage{bookmark}
\input{glyphtounicode}
\usepackage[
  backend=biber,
  natbib=true,
  style=ieee,
  citestyle=numeric-comp,
  giveninits=true,
  minbibnames=1,
  maxcitenames=2,
  mincitenames=1,
  sorting=none,
  maxbibnames=6,
  doi=false,
  isbn=false,
  url=false,
  eprint=false
]{biblatex}

\DeclareFieldFormat[article]{eid}{Art.~no.\space#1}

\usepackage{siunitx}
\usepackage[capitalize,nameinlink]{cleveref}
\crefname{equation}{Eq.}{Eqs.}
\Crefname{equation}{Equation}{Equations}
\crefname{figure}{Fig.}{Figs.}
\Crefname{figure}{Figure}{Figures}
\crefname{section}{Section}{Sections}
\crefname{subsection}{Section}{Sections}

\usepackage[printonlyused]{acronym}
\acrodef{LHD}{load-haul-dump}
\acrodef{ICR}{instantaneous center of rotation}
\acrodefplural{ICR}[ICRs]{instantaneous centers of rotation}
\acrodef{IMU}{inertial measurement unit}
\acrodef{ICP}{iterative closest point}
\acrodef{RPE}{relative pose error}
\acrodefplural{RPE}[RPE]{relative pose errors}
\acrodef{IQR}{interquartile range}
\acrodef{PCK}{point-contact kinematic model}
\acrodef{FSQ}{finite-support quadratic model}
\acrodef{STCA}{single-track center-articulated}

\usepackage{tabularx}
\usepackage{makecell}

\DeclareMathAlphabet{\mathcal}{OMS}{ztmcm}{m}{n}

\newcommand{\SE}{\mathrm{SE}}

\newcommand{\se}{\mathfrak{se}}
\newcommand{\Ad}{\mathrm{Ad}}
\newcommand{\dd}{\mathrm{d}}
\newcommand{\R}{\mathbb{R}}
\newcommand{\Sph}{\mathbb{S}}

\DeclareMathOperator*{\argmin}{arg\,min}

%% file: metadata.tex
\newcommand{\PaperTitle}{Steering Through Contact: A Finite-Support Motion Model for Single-Track Center-Articulated Robots}
\hypersetup{
  pdftitle={\PaperTitle},
  pdfauthor={Mohamed Dhia Ounally, Nicolas Samson, Mathis Turgeon-Roy, Veronica Vannini, Johann Laconte, Fran\c{c}ois Pomerleau},
  pdfsubject={Finite-support quadratic motion prediction for single-track center-articulated field robots},
  pdfkeywords={field robotics, articulated tracked robots, contact modeling, finite-support quadratic model, point-contact kinematics, lateral slip, motion prediction},
  pdflang={en-US},
  keeppdfinfo=true,
  pdfdisplaydoctitle=true,
  bookmarksnumbered=true,
  bookmarksopen=false
}

%% file: authors.tex
\author{\normalsize Mohamed Dhia Ounally$^{1,2}$, Nicolas Samson$^{1}$, Mathis Turgeon-Roy$^{1}$,\\[-1pt]
\normalsize Veronica Vannini$^{1}$, Johann Laconte$^{2}$, and Fran\c{c}ois Pomerleau$^{1}$%
\thanks{$^{1}$Norlab, Universit\'e Laval, Qu\'ebec, QC, Canada.}%
\thanks{$^{2}$TSCF, INRAE, Clermont-Ferrand, France.
\href{mailto:mohamed.ounally@inrae.fr}{mohamed.ounally@inrae.fr}.}}

%% file: acknowledgments.tex
\section*{Acknowledgment}
Supported by ANR (Johann Laconte's Junior Research Chair), I-SITE CAP 20-25 (``Innovation Transportation and Production Systems''), and IID, Universit\'e Laval (Excellence Scholarships to Mohamed Dhia Ounally).